\pdfoutput=1
\documentclass[11pt]{article}

\usepackage{emnlp2021}

\usepackage{times}
\usepackage{latexsym}

\usepackage{tabulary}
\usepackage{amsmath}
\usepackage{amssymb}
\usepackage{MnSymbol}%
\usepackage{wasysym}%
\usepackage{tikz}
\usepackage{pgfplots}
\usepgfplotslibrary{groupplots}
\usetikzlibrary{calc}
\usepackage{xcolor}
\usepackage{color, colortbl}
\usepackage[normalem]{ulem}
\usepackage{url}
\usepackage{multirow}
\usepackage{float}
\usepackage{paralist}
\usepackage{comment}
\pgfplotsset{compat=newest}

\usepackage{color}
\usepackage{colortbl,array,xcolor}
\usepackage{silence}

\usepackage{multicol}
\usepackage{multirow}
\usepackage{booktabs}
\usepackage{soul}
\usepackage{pgfplots}
\usepgfplotslibrary{groupplots}
\usetikzlibrary{matrix}

\newcolumntype{?}{!{\vrule width 1pt}}

\def\similarity{\mathrm{sim}}

\def\vphi{{\boldsymbol{\phi}}}

\def\vb{{\mathbf{b}}}

\def\vk{{\mathbf{k}}}

\def\vq{{\mathbf{q}}}

\def\vv{{\mathbf{v}}}

\def\vx{{\mathbf{x}}}
\def\vy{{\mathbf{y}}}
\def\vz{{\mathbf{z}}}

\def\mS{{\mathbf{S}}}

\def\mW{{\mathbf{W}}}

\def\gO{{\mathcal{O}}}

\newcommand{\src}{{\mathrm{src}}}
\newcommand{\tgt}{{\mathrm{tgt}}}

\newcommand{\out}{{\mathrm{out}}}

\newcommand{\relu}{{\operatorname{relu}}}
\newcommand{\elu}{{\operatorname{elu}}}

\newcommand{\interalia}[1]{\citep[\emph{inter alia}]{#1}}

\usepackage{mathtools}
\usepackage{arydshln}

\usepackage{graphicx}  %Required
\graphicspath{{images/}{plots/}}
\usepackage{pgfplots}
\pgfplotsset{compat=1.12}

\usepackage{nccmath}

\definecolor{cyellow}{HTML}{CC79A7}
\definecolor{nred}{HTML}{D55E00}
\definecolor{nblue}{HTML}{0072B2}
\definecolor{ngreen}{HTML}{009E73}
\definecolor{ggreen}{HTML}{82B366}
\definecolor{bblue}{HTML}{6C8EBF}

\usepackage{xspace}

\newcommand{\TRNN}{T2R\xspace}

\usepackage[T1]{fontenc}
\usepackage[utf8]{inputenc}

\usepackage{microtype}

\title{Finetuning Pretrained Transformers into RNNs}

\author{
    Jungo Kasai$^{\heartsuit}$\thanks{\ \ Work was done during an internship at Microsoft.}  
\quad
\quad 
\textbf{Hao Peng}$^{\heartsuit}$
\quad 
\quad 
\textbf{Yizhe Zhang}$^{\clubsuit}$
\quad 
\textbf{Dani Yogatama}$^{\spadesuit}$\\
\textbf{Gabriel Ilharco}$^{\heartsuit}$ 
\quad
\textbf{Nikolaos Pappas}$^{\heartsuit}$ 
\quad 
\textbf{Yi Mao}$^{\clubsuit}$
\quad 
\textbf{Weizhu Chen}$^{\clubsuit}$
\quad
	\textbf{Noah A.\ Smith}$^{\heartsuit\diamondsuit}$\\
\\
    $^{\heartsuit}$Paul G.\ Allen School of Computer Science \& Engineering, University of Washington
    \\
    $^{\clubsuit}$Microsoft
\quad $^{\spadesuit}$DeepMind
\quad $^{\diamondsuit}$Allen Institute for AI\\
    {\tt \{jkasai,hapeng,gamaga,npappas,nasmith\}@cs.washington.edu}\\
    {\tt \{Yizhe.Zhang, maoyi, wzchen\}@microsoft.com}\\
    {\tt dyogatama@google.com}
}

\begin{document}
\setlength{\abovedisplayskip}{4pt}
\setlength{\belowdisplayskip}{4pt}
\setlength{\abovedisplayshortskip}{4pt}
\setlength{\belowdisplayshortskip}{4pt}
\maketitle
\begin{abstract}
Transformers have outperformed recurrent neural networks (RNNs) in natural language generation.
But this comes with a significant computational cost, as the attention mechanism's complexity scales quadratically with sequence length.
Efficient transformer variants have received increasing interest in recent works. 
Among them, a linear-complexity \emph{recurrent} variant has proven well suited for autoregressive generation.
It approximates the softmax attention with randomized or heuristic feature maps,
but can be difficult to train and may yield suboptimal accuracy.
This work aims to \emph{convert} a pretrained transformer into its efficient recurrent counterpart,
improving efficiency while maintaining  accuracy.
Specifically, we propose a \textit{swap-then-finetune} procedure:
in an off-the-shelf pretrained transformer, 
we replace the softmax attention with its linear-complexity recurrent alternative
and then finetune.
With a learned feature map, our approach 
provides an improved tradeoff between efficiency and accuracy over the standard transformer and other recurrent variants.
We also show that the finetuning process has lower training cost relative to training these recurrent variants from scratch.
As many models for natural language tasks are increasingly dependent on large-scale pretrained transformers, this work presents a viable approach to improving inference efficiency without repeating the expensive pretraining process.\footnote{\url{https://github.com/jungokasai/T2R/}.}

\end{abstract}

\section{Introduction}
Transformer models \cite{Vaswani2017AttentionIA} have advanced the state of the art beyond recurrent neural network models (e.g., LSTMs, \citealp{hochreiter1997long}; GRUs, \citealp{cho-etal-2014-properties}) across a wide range of natural language processing tasks.
In particular, the transformer architecture has been widely used in autoregressive modeling such as language modeling \cite{Baevski2019AdaptiveIR} and machine translation \cite{Vaswani2017AttentionIA}.
%such as machine translation \cite{Vaswani2017AttentionIA} and language modeling \cite{Baevski2019AdaptiveIR}.
The transformer makes crucial use of interactions between feature vectors over the input sequence through the attention mechanism \citep{Bahdanau2014NeuralMT}.
However, this comes with significant computation and memory footprint during generation.
%If the input sequence is predetermined, the time complexity can be reduced with parallel computing over input words on modern GPUs or TPUs \cite{deepshallow}.
Since the output is incrementally predicted 
% one by one 
conditioned on the prefix, generation steps cannot be parallelized over time steps
and require quadratic time complexity in sequence length.
The memory consumption in every generation step also grows linearly as the sequence becomes longer.
This bottleneck for long sequence generation limits the use of large-scale pretrained transformers, such as GPT-3 \cite{gpt3}, Image Transformer \cite{imagetransformer}, and DALL-E \cite{DALLE}.
%On the other hand, this increased complexity implies substantial overhead in autoregressive generation where words are predicted one by one conditioned on all previous words \cite{Gu2017NonAutoregressiveNM}. 

Recent work aims at reducing 
the overhead
% this time and memory overhead 
of autoregressive transformers \interalia{Child2019, Kitaev2020ReformerTE, longformer}. 
%While those alternatives are able to achieve comparable performance to the standard transformer architecture in some tasks, their applicability is often more limited \cite{Tay2020EfficientTA}.
%\hao{reworded}
% than the \yz{vanilla?} transformer. 
%For example, Linformer \cite{Wang2020LinformerSW} projects down the sequence length dimension from the number of input words to a fixed size, but this design assumes the sequence length is invariant\yz{...sequence length needs to be padded to the same length, rendering additional computation cost for short sequences}; \textit{autoregressive} modeling where each output token is predicted by its prefix cannot be implemented in a straightforward way.\yz{maybe use ``1)'', ``2)'' rather than ``;'' to separate the two reasons?}
%\citet{katharopoulos-et-al-2020} illustrated that their attention with linear complexity can achieve competitive performance in autoregressive image generation on vision benchmarks, but their model yields much degraded accuracy on natural language tasks, such as language modeling and machine translation, compared to the transformer architecture \cite{RFA}. \gamaga{"much degraded accuracy is reading a bit weird for me, but not exactly sure why"}\yz{any reason to mention about why transformers as RNN did bad for NLP tasks, that leads to our remedy,``converter'', to their approaches? Need to smooth the logic flow here. }
%\hao{suggestion: talk a bit more about these recent methods, especially those we build on before we contrast our method to them}
Among them are recurrent alternatives that approximate the standard softmax attention \cite{katharopoulos-et-al-2020, RFA, performer,schlag2021linear}.
Similar to recurrent neural networks (RNNs), those models represent the context by a recurrent state with a fixed size, thereby achieving linear time and constant memory complexity in generation sequence length.
When the recurrent state size is smaller than the sequence length, these variants provide substantial speed and memory advantages over the transformer.
A small state size, however, tends to deteriorate the generation quality \cite{RFA}, leading to a tradeoff between efficiency and accuracy.

This work improves the balance between efficiency and accuracy by a \textit{conversion} approach: instead of training a recurrent alternative from scratch, we develop a method to \textit{convert} a pretrained transformer into an efficient RNN that speeds up generation and reduces memory footprints.
%\hao{suggestion: be more specific about RNN's decoding efficiency advantage. something like this would do: although it is outperformed by transformers in accuracy, it has the efficiency advantage in settings like decoding}
Our conversion proceeds with a \textit{swap-then-finetune} process.
Specifically, we change the exponential similarity function in the attention mechanism to the dot product after a single-layer MLP feature mapping.
We then finetune the MLP parameters and the other network parameters.
Our experiments in language modeling and machine translation show that the conversion can compress the context into a much smaller recurrent state than the sequence length (e.g., 1/16 of the sequence length in WikiText-103 language modeling) while retaining high accuracy.
%For example, in the language modeling task
%\hao{it doesn't come through how the size of recurrence state might have an impact on efficiency. suggestion: earlier this paragraph we can talk more about these recurrent attention works, and be very specific about the accuracy vs. efficiency trade-off there.
%Something like this would do: e.g., RFA/performer's recurrence size can be much larger than the sequence lengths in tasks involving working with moderate-length inputs, limiting their efficiency gain in such settings.
%And it can more catchy if we compare the actual numbers: 256 vs. 32 or 16, 1/8 or 1/16 of the size, same accuracy, more efficiency gain.
%Anyways, these small sizes are very surprising and exciting, and we should make them so!
%}
In addition, this conversion requires much less GPU time than training randomly initialized models from scratch.
%and improves the balance between efficiency and accuracy over the recent autoregressive transformer variants with lightweight attention.
%retains high accuracy with a substantial speedup in autoregressive generation. 
%substantially outperforms approaches that train new networks from random initialization.
%and narrows the accuracy gap from the transformer.

%Our conversion approach relies on a pipeline: we first train a transformer model, then convert it into an efficient recurrent neural network. \hao{
%this implies additional training overhead.
%suggestion: we can argue that pretrained transformers
%are everywhere. we just finetune them into RNNs, and imply that the pretraining is not part of the overhead.
%oh you actually say this later. I suggest dropping this pipeline thing then.
%}
%By design, the first step of training a transformer remains as necessary overhead.
State-of-the-art models in many natural language tasks are increasingly dependent on large-scale pretrained transformer models (e.g., GPT-2, \citealp{gpt2}; BERT, \citealp{devlins2019bert}; RoBERTa, \citealp{Liu2019RoBERTaAR}; T5, \citealp{2020t5}; BART, \citealp{lewis-etal-2020-bart}; DeBERTa, \citealp{he2021deberta}). 
Converting a large off-the-shelf transformer to a lightweight inference model without repeating the whole training procedure is particularly useful in many downstream applications. 
Our work focuses on text generation and presents a viable approach towards efficient inference with high accuracy.
%\footnote{Our code will be available at \url{https://github.com/jungokasai/T2R}.}

%Our conversion approach provides a way to benefit from off-the-shelf models without training a large model from random initialization.
%instead of training new models with a different network architecture from scratch.
%training new models with different network architectures from random initialization is infeasible in many scenarios.

\section{Convert a Transformer into an RNN}
\label{T2RNN}
%\hao{we need to clarify why we don't compare to Performer at an important place.}
%\hao{suggestion: include a detailed decoding complexity analysis of the three kernel variants as the softmax baseline.
%This would help distinguish MLP from RFA/Performer and ELU, and highlight the technique that merges two projections at inference time}
%\hao{another thing we need to clarify: finetuning pretrained transformer into some efficienct variants can be used
%for many different models beyond these three RNN-ish ones.
%it would be great if we can find something concrete arguing that there is something unique about finetuning these kernel-based models.
%}
The transformer architecture consists of multihead attention, feedforward, and layer normalization modules \cite{Vaswani2017AttentionIA}.
% With a quadratic complexity in length, the attention modules are usually the bottleneck in autoregressive decoding. \hao{reworded}
% Multihead attention modules cost quadratic time complexity in input sequence length, while the others are linear.
%If the input sequence is fully revealed, the attention computation can be parallelized over the sequence.
When a transformer is \emph{trained} for a sequence generation task with teacher forcing \citep{teacher-forcing}, 
the attention can be parallelized over positions 
because the target sequence is fully available.
During \emph{generation}, on the other hand, the output is incrementally constructed.
As a result, the attention becomes an inference bottleneck for long sequences. % as the generation output gets longer.
We present a method to eliminate this bottleneck by converting a pretrained transformer into an efficient RNN of linear time and constant space complexity.
We provide a detailed complexity analysis in terms of the sequence length and model dimensions.
%The transformer architecture \cite{Vaswani2017AttentionIA} differs from recurrent neural networks such as LSTMs \cite{hochreiter1997long} and GRUs \cite{cho-etal-2014-properties} in its parallel structure. 
%Here we review the architecture and discuss its implications in training and inference.
%\label{sec:arch}

\subsection{Multihead Attention}
%\hao{nit: maybe promise the readers that we choose to be very articulated on the model dimensions,
%since they are crucial to a complexity analysis that comes later?}
The attention module takes as input sequences of \textit{source} and \textit{target} vectors.
The source vectors are used to produce \textit{key} and \textit{value} features, while the target vectors are mapped to \textit{query} vectors.
%Different types of attention involve different combinations of source and target vectors.
%In the encoder-to-decoder (cross) attention of a sequence-to-sequence model, the source vectors consist of encoder representations and target vectors come from the decoder's input.
%In  self and causal attention, the source and target come from the same input, while the encoder-to-decoder (cross) attention in sequence-to-sequence models have different source and target vectors.
More formally, denote by $\{ \vx_{i}^{\tgt}\}_{i=1}^N$ and $\{\vx_{j}^{\src}\}_{j=1}^M$
the target and source vectors,
% Suppose that the input sequences of target and source vectors are $\left \{ \vx_{i}^{\tgt}\right\}_{i=1}^N$ and $\left\{\vx_{j}^{\src}\right\}_{j=1}^M$ respectively 
where $\vx_{i}^{\tgt}, \vx_{j}^{\src} \in \mathbb{R}^{h}$ and $h$ is the model dimensionality.
We assume $r$ attention heads of $d$ dimensions ($h = dr$).
For each head, the input vectors are first mapped to $d$ dimensional \textit{query}, \textit{key}, and \textit{value} features by learned affine transformations with $\mW_\ast \in \mathbb{R}^{d \times h}$ and $\vb_{\ast} \in \mathbb{R}^{d}$:
%\hao{nit: suggest dropping the bias terms, unless they are important to some later discussion}
%\jungo{I guess clarity/simplicity vs.\ precision here. I'm leaning toward keeping bias terms, but we will see.}
\begin{subequations}
\begin{align}
\vq_{i} &= \mW_{q} \vx^{\tgt}_i + \vb_q, \label{eq:q}\\
\vk_{j} &=  \mW_{k} \vx^{\src}_j +\vb_k, \quad
\vv_{j} = \mW_{v}\vx^{\src}_j + \vb_v. 
\label{eq:kv}
\end{align}
\end{subequations}
The similarities of each query vector $\vq_i$ with all $M$ key vectors are computed and normalized to produce attention coefficients, which are then used to output a weighted average of the value vectors \cite{Vaswani2017AttentionIA}:
\begin{subequations}
\begin{align}
\vx^{\out}_i  &=  \sum_{j=1}^M \frac{\similarity\left(\vq_i, \vk_j\right)}{\sum_{\ell=1}^M \similarity\left(\vq_i, \vk_{\ell}\right)} \vv_j,\label{eq:attn}\\
%\end{align}
%where
%\begin{align}
%\text{where} 
\similarity (\vx, \vy)&=
    %\exp\left(\frac{\vx^\top \vy}{\sqrt{d}}\right) 
    \exp\left(\vx \cdot \vy/\sqrt{d}\right).\label{eq:dot_exp}
\end{align}
\end{subequations}
Multihead attention runs this procedure for each of the $r$ heads in parallel and concatenates $r$ output vectors %of $d$ dimensions 
to get the final $h$ dimensional vector.\footnote{Layer normalization \cite{Ba2016LayerN}, residual connection \cite{resnet}, and projection are suppressed for brevity.}  
\begin{figure*}[t]
\centering
    \includegraphics[width=0.90\textwidth]{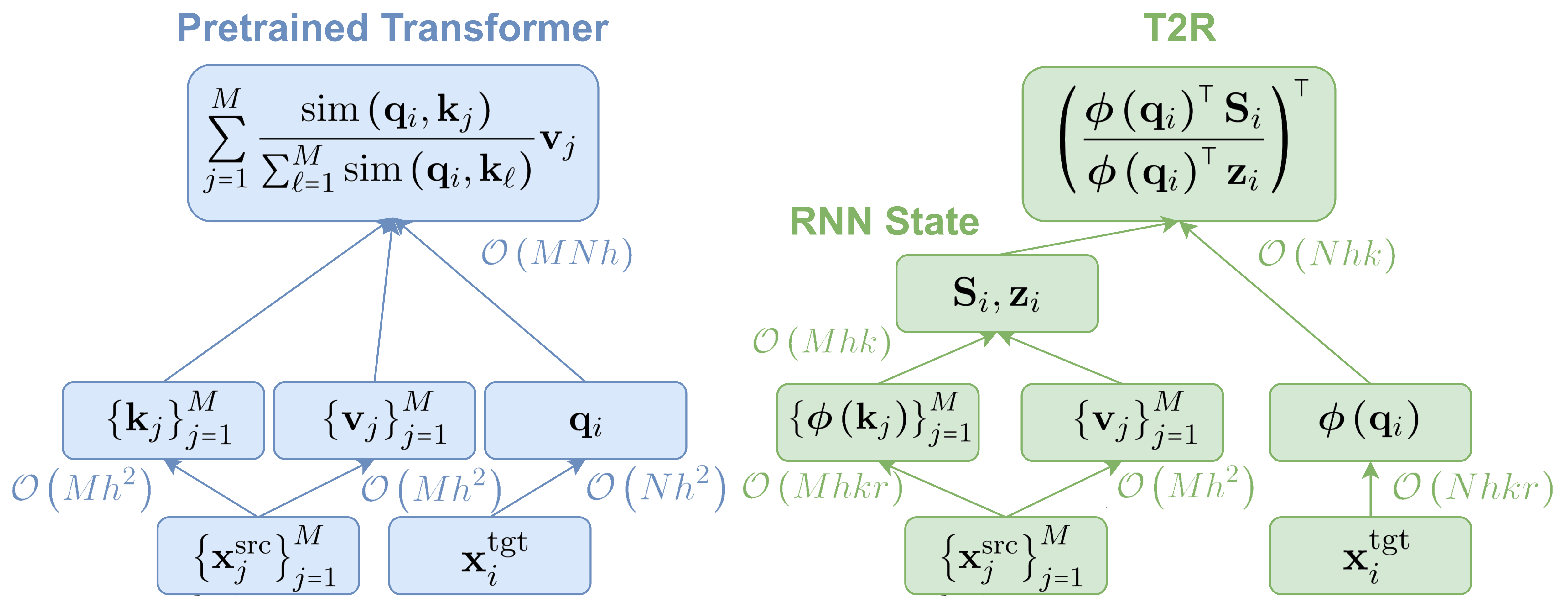}
\caption{Attention computation steps and their time complexity in pretrained transformer and \TRNN models during inference generation. 
Features  $\vphi(\vq_i)$ and $\vphi(\vk_j)$ are directly computed from input vectors,
and $\vq_i$ and $\vk_j$ are never constructed.
$M$: source length; $N$: target length; $h$: model dimensions; $k$: feature size; $r$: \# heads.
%\yz{Mhk/Nhk to Mhkr/Nhkr for the right panel. I would also indicate the dimension of each node}
}
\label{fig:t2rnn}
%\vspace{-0.4cm}
\end{figure*}

\paragraph{Generation Speed Overhead}
Fig.\ \ref{fig:t2rnn} depicts the transformer computation steps from input vectors and their time complexity.
We assume that the time complexity of multiplying an $n \times m$ matrix by an $m \times k$ is $\gO(nmk)$ as implemented in cuBLAS \cite{cublas}.\footnote{If the batch size is small enough, parallelization can speed up matrix multiplication.}
It consists of the following two stages.
\begin{compactitem}
\item \textbf{Feature Mapping}: computation of $\{\vq_i\}_{i=1}^N$,  $\{\vk_j\}_{j=1}^M$, and $\{\vv_j\}_{j=1}^M$ for all $r$ heads from input vectors (Eqs.\ \ref{eq:q}-\ref{eq:kv}).
Time complexity of $\gO(Nh^2)$, $\gO(Mh^2)$, and $\gO(Mh^2)$.
\item \textbf{Attention}: weighted average over the value vectors (Eq.\ \ref{eq:attn}). $\gO(MNh)$, quadratic in sequence length ($M$, $N$).
\end{compactitem}
\paragraph{Generation Memory Overhead}
In autoregressive generation, 
%output tokens are predicted incrementally. \hao{A bit repetitive}
%In this case, 
query, key, and value vectors consume space complexity of $\gO(h)$, $\gO(Mh)$, and $\gO(Mh)$ in every generation step.
Every step's attention weight (Eq.~\ref{eq:attn}) spans over $M$ source positions, taking $\gO(Mr)$ space, linear in sequence length $M$.

%during training, autoregressive generation \cite{deepshallow} 

\subsection{Converting Transformers to RNNs}
\label{sec:convert}
To address this generation bottleneck of quadratic time and linear space, we propose Transformer-to-RNN (\TRNN), a method to convert a pretrained transformer to an RNN inference model of linear time and constant memory complexity in sequence length (Fig.\ \ref{fig:t2rnn}).
\TRNN follows a swap-then-finetune procedure that modifies the attention computation of a pretrained transformer, and finetunes the model with the task objective.

%\hao{again, maybe consider dropping bias term}
We first replace the dot-then-exponential similarity function in a pretrained transformer (Eq.~\ref{eq:dot_exp}) by
\begin{subequations}
\vspace{-0.5cm}
\begin{align}
\widetilde{\similarity} \left( \vx, \vy\right) 
&= \vphi\left(\vx\right)\cdot \vphi\left(\vy\right),\\
%\end{align}
%\begin{align}
%\text{where} \quad 
\vphi \left (\vx \right) &=  \relu\left(\mW_{\vphi} \vx + \vb_{\vphi}\right).
\end{align}
\end{subequations}
Here $\mW_\vphi \in \mathbb{R}^{k \times d}$ and $\vb_\vphi \in \mathbb{R}^{k}$ are 
learned parameters
of a single-layer MLP.
They map a $d$ dimensional vector to a $k$ dimensional kernel feature space.
The $\relu$ activation \cite{Fukushima1980} ensures that the features are non-negative.\footnote{We found that $\relu$ stabilized training by prohibiting
negative similarities $\vphi(\vq)\cdot \vphi(\vk)$. Other activation functions, such as $\mathrm{cos}$, $\mathrm{tanh}$, and $\mathrm{elu}$, did not improve performance.} %so that the attention weight yields a valid probability distribution over the input sequence
%\hao{relu doesn't guarantee a distribution since it can produce all-zero attention weights?
%do we need to address this?}
% Without it training diverges.
% We found that removing the $\relu$ activation indeed leads to training divergence.
Different MLP parameters are used for different attention heads, and thus we add a total of $rk(d+1)$ learnable parameters per layer (less than 0.2\% parameter increase in our language model, \S\ref{sec:experiments}).
We then finetune all parameters in this modified network, including the MLP parameters, with the original task objective.\footnote{We tried training the MLP parameters only, but this setting resulted in degraded development performance.}

%Using this modified similarity function,
During inference generation, we reformulate the attention computation (Eq.~\ref{eq:attn}) as
\begin{align}
\begin{split}
\widetilde{\vx}^{\out}_i  &=  
\sum_{j=1}^M \frac{\widetilde{\similarity}\left(\vq_i, \vk_j\right)}{\sum_{\ell=1}^M \widetilde{\similarity}\left(\vq_i, \vk_{\ell}\right)} \vv_j \\
%&=\sum_{j=1}^M \frac{\vphi\left(\vq_i\right)\cdot \vphi\left(\vk_j\right)}{\sum_{\ell=1}^M \vphi\left(\vq_i\right)\cdot \vphi\left(\vk_\ell\right)} \vv_j \\
&= \left(   \frac{\vphi\left(\vq_i\right)\cdot \sum_{j=1}^M \vphi\left(\vk_j\right) \otimes \vv_j}{ \vphi\left(\vq_i\right)\cdot \sum_{\ell=1}^M \vphi\left(\vk_\ell\right)} \right)^\top
\end{split}
\end{align}
by the associativity of matrix multiplication. % \cite{katharopoulos-et-al-2020}.
%The softmax attention corresponds to a case where $\vphi_{\softm} \left( \vx \right) \in \mathbb{R}^{\infty}$ (similar to the Gaussian kernel), but when $\vphi\left(\vx\right)$ is finite dimensional, the last formulation allows us to reduce the complexity to linear with respect to the sequence lengths.
This formulation lends itself to recurrent computation.
In causal attention where each query only attends to its prefix to predict the next word ($M=i$), define states:
\begin{align}
\mS_i = \sum_{j=1}^i \vphi \left( \vk_j \right) \otimes  \vv_j, \quad
\vz_i   = \sum_{j=1}^i \vphi \left(\vk_j \right) \label{eq:sz}
\end{align}
where $\mS_{i}, \vz_{i} \in \mathbb{R}^{k \times d}, \mathbb{R}^{k}$.
These states can be computed recurrently \cite{katharopoulos-et-al-2020}:
\begin{align}
\mS_{i} = \mS_{i-1} + \vphi \left( \vk_i \right) \vv_i^\top  \quad \vz_{i} = \vz_{i-1} + \vphi \left(\vk_i \right) \label{eq:recurrent}
\end{align}
In the self-attention or encoder-to-decoder (cross) attention of a sequence-to-sequence model, $\mS_{i}$ and $\vz_{i}$ are constant with respect to $i$ and only need to be computed once.
Given the two states at position $i$, we can obtain the output vector: 
\begin{align}
\widetilde{\vx}_{i}^{\out} = \left(\frac{\vphi\left(\vq_i\right)^\top \mS_i}{\vphi\left(\vq_i\right)^\top \vz_i}\right)^\top \label{eq:out}
\end{align}
This avoids quadratic computation with respect to the input sequence length.
%We then finetune this RNN model, including the MLP parameters, with the task objective.
We also speed up inference by merging the MLP feature map with the affine feature maps that produce queries and keys.
\begin{subequations}
\begin{align}
  %\vphi \left( \vq_{i} \right) &= \left(\mW_\phi \mW_q\right) \vx^\tgt_i+ \left(\vb_\phi + \mW_\phi \vb_q  \right)\label{eq:phiq}\\
  \vphi \left( \vq_{i} \right) = \relu\left( \widetilde{\mW}_q \vx^\tgt_i + \widetilde{\vb}_q\right), \label{eq:phiq} \\
  \vphi \left( \vk_{j} \right) = \relu\left(\widetilde{\mW}_k \vx^\src_j + \widetilde{\vb}_k\right),\label{eq:phik}
  %\vphi \left( \vk_{j} \right) &= \left(\mW_\phi \mW_k\right) \vx^{\src}_j+ \left(\vb_\phi + \mW_\phi \vb_k  \right)\label{eq:phik}
\end{align}
\begin{align}
\text{where} \quad \widetilde{\mW}_q = \mW_\phi \mW_q, \quad \widetilde{\mW}_k = \mW_\phi \mW_k, \label{eq:merge_q}\\
  \widetilde{\vb}_q = \vb_\phi + \mW_\phi \vb_q, \quad \widetilde{\vb}_k = \vb_\phi + \mW_\phi \vb_k. \label{eq:merge_k}
\end{align}
\end{subequations}%
%\yz{explicitly state that only $\mW_{\phi}, \vb_\phi$ need to be optimized?}
After the model is trained, Eqs.~\ref{eq:merge_q}--\ref{eq:merge_k} are computed once before generation; 
the intermediate features of $\vq_i$ and $\vk_j$ are never computed during inference. 

%\hao{need to be more specific that $\mW_\phi \mW_q$ is computed ``offline'' once after the model is trained,
%and we never recompute it for each inference batch.
%could also stress that this reduces two matrix-vector mult (as in previous works) to only one,
%and that if we use small feature map, this is even cheaper than standard multihead attn.}
%results in time and space complexity of $\gO(Nk)$, linear in input sequence length. 

\paragraph{Generation Speed Overhead}
The time complexity of each step in a \TRNN model is shown in Fig.\ \ref{fig:t2rnn}.
Similar to the transformer, it proceeds over two stages.
\begin{compactitem}
  \item \textbf{Feature Mapping}: computation of $\{\vphi(\vq_i )\}_{i=1}^N$, $\{\vphi(\vk_j)\}_{j=1}^M$, and $\{\vv_j\}_{j=1}^M$ for all $r$ heads (Eqs.\ \ref{eq:phiq}--\ref{eq:phik}). 
  Time complexity of $\gO( Nhkr)$, $\gO( Mhkr )$, and $\gO( Mh^2)$.
  %\hao{It seems that we make some assumption on matrix-vector mult's time complexity. If so we need to specify and justify it.}
  %\jungo{Hmmn, I don't think so? nxk, kxl should be nkl complexity.}
  \item \textbf{Attention}: the RNN states and the outputs for $r$ heads (Eqs.\ \ref{eq:sz}--\ref{eq:out}) are computed with $\gO(Mhk)$ and $\gO (Nhk)$.
  %\hao{ditto}
\end{compactitem}
Comparing this with the pretrained transformer, we see that if the feature size is much smaller than input sequence lengths ($k\ll M,N$), the change in the attention stage from $\gO(MNh)$ to $\gO(hk(M+N))$ in \TRNN brings a substantial speedup.
%the \TRNN model has the following two speed advantages.
%\begin{compactitem}
%    \item If the feature size is much smaller than input sequence lengths ($k<<M,N$), the conversion of the attention stage from $\gO(MNh)$ to $\gO(Mhk)+\gO(Nhk)$ in \TRNN brings a substantial speedup.
%    %Later we will see a tradeoff between speed and accuracy: a smaller $k$ speeds up generation at the expense of an accuracy drop.
%    \item If the feature size is smaller than the head dimensions ($k<d$), feature mapping has a smaller computational cost since $kr<dr=h$.
%  \end{compactitem}
\paragraph{Generation Memory Overhead}
\TRNN only needs to store the RNN state, and thus its space complexity is $\gO(hk)$, constant in sequence length. 
This implies reduction in memory footprint when $k\ll M$, compared to the transformer's $\gO(Mh)$.
%This means that if the kernel size $k<<M,N$

\subsection{Autoregressive Linear Transformers}
\label{sec:linear_transformers}
%\hao{this sections reads like a defensive list of points differentiating our model from previous works.
%we can be firm about the advantages of MLP feature map.
%Also, some of the discussion of previous work is too detailed and fits better in related work.
%}
%\jungo{Shrunk the paragraphs and removed details.}
In principle, any kernel function can be used as the similarity function in Eq.\ \ref{eq:attn} \cite{tsai-etal-2019-transformer}.
Previous work proposed several untrainable feature map functions $\vphi$ and developed autoregressive transformer variants with linear time and constant space complexity in sequence length \cite{katharopoulos-et-al-2020, RFA, performer}.
%Performer \cite{performer} also applied similar random approximation with a positive constraint.
%and evaluated the linear transformer on protein sequence modeling and other long sequence benchmarks.
%Similar to those prior works, it is also possible to train our \TRNN model from random initialization.
%However, we will show in a later section that the \TRNN conversion %is crucial to achieve high accuracy especially when the feature size is small.
%\hao{this is a good place to summarize the advantages of MLP feature map and contrast it against others.}
While those models follow similar computation steps to \TRNN, there are several differences in generation efficiency.
Since the feature map in \citet{katharopoulos-et-al-2020} preserves input dimensions, the feature size is always the same as the head dimensions ($k=d$).
This means that the speedup and memory savings from using a small feature size are restricted by design.
%(\S\ref{sec:results}).
In our experiments (\S\ref{sec:results}), our \TRNN models gain further efficiency by using a feature size that is even smaller than the head dimensions ($k=32$ and $d=128$ for language modeling).
\citet{RFA} and \citet{performer} scale query and key vectors by their norms before the random approximation to bound the error.
Consequently, the feature mapping stage needs additional steps of producing intermediate $\vq$ and $\vk$ and scaling them.
\TRNN suppresses these steps and speeds up generation further (\S\ref{sec:results}).

\section{Experiments}
\label{sec:experiments}
We present extensive experiments on standard benchmarks for language modeling and machine translation.
Our results show that \TRNN achieves efficient autoregressive generation while retaining high accuracy.

\subsection{Baselines and Comparison}
We compare performance with previous transformer models for autoregressive generation with linear time and constant space complexity in input sequence length.\footnote{See \S\ref{sec:related_work} for our discussion on more transformer variants with linear time complexity, but most of those variants need modifications for autoregressive modeling and have yet to be empirically evaluated in autoregressive generation tasks.} 
As discussed in \S\ref{sec:linear_transformers}, those prior methods correspond to two different untrainable feature maps $\vphi$.
We experiment with two types of feature maps for comparisons: \textbf{ELU} ($\vphi\left(\vx\right) = \elu \left(\vx\right) + 1$, \citealp{katharopoulos-et-al-2020}); \textbf{RFA} (random feature approximation with softmax temperature reparameterization, \citealp{RFA}).
Each feature map is evaluated in two settings: random initialization and pretrain.
Random initialization is our reimplementation of the experiments in \citet{katharopoulos-et-al-2020} and \citet{RFA}.
The pretrain setting follows the same protocol as \TRNN except that we use different feature maps $\vphi$ than our proposed one-layer MLP with $\relu$ activation.
Positive orthogonal random features (\textbf{Performer}, \citealp{performer}) provide similar random approximation to RFA and were evaluated in the biology domain, but we found that this method caused training divergence in the language modeling task.\footnote{Our implementation closely follows the code released by the authors (\url{https://github.com/lucidrains/performer-pytorch/blob/main/performer_pytorch/performer_pytorch.py\#L75-L81}), but does \textit{not} subtract the maximum logit; otherwise it would disallow the linear complexity in causal attention. We conjecture that this is the reason why Performer becomes less stable in our experiments. We suspect that some techniques are necessary to improve numerical stability in language modeling and machine translation.
}
%Note that by design, the kernel size $k$ for ELU has to be the same as the head dimension whereas RFA and MLP allow for a smaller dimension to further reduce generation time and memory overhead (\S\ref{sec:results}).

%\paragraph{Speed Measurement}
%For machine translation, we \cite{RFA} 
%evaluation and generation in language modeling, we implement a custom cuda kernel on a RTX GPU.

\subsection{Setup and Implementations}
We apply our method to causal attention in language models and both cross and causal attention in machine translation.
For language modeling, we use a 32-dimensional feature map function.
We do not modify the encoder in machine translation as its generation speed overhead is much less significant than the decoder \cite{deepshallow}.
Our exploration showed that reducing the feature size of causal attention tends to have less impact on the final translation accuracy as opposed to cross attention; we use feature sizes of 32 and 4 for cross and causal attention, respectively. 
This observation is consistent with previous work that showed that causal attention can be more drastically simplified than cross attention in transformer machine translation models \cite{you-etal-2020-hard, synthesizer}.

%For both language modeling and machine translation, we base our model and optimization hyperparameters on previous strong models \cite{Baevski2019AdaptiveIR, Vaswani2017AttentionIA}.
%We found that some hyperparameters, such as the learning rate for the \TRNN finetuning, need to be different from those for randomly initialized training (see Appendix \ref{sec:hyper}). 
%All models are implemented using \texttt{fairseq} \cite{ott-etal-2019-fairseq}.
%and trained on 8 NVIDIA V100 GPUs.
%We apply mixed-precision training \cite{micikevicius2018mixed, ott-etal-2019-fairseq} to speed up training.

\subsubsection{Language Modeling}
We use the WikiText-103 benchmark, which consists of 103M tokens sampled from English Wikipedia \cite{wiki103}.
We choose similar hyperparameters to prior work \cite{Baevski2019AdaptiveIR, layerdrop}: 32 layers, 8 heads, 128 head dimensions, 1024 model dimensions, 4096 fully connected dimensions and dropout \cite{dropout} and layer dropout rates of 0.2.
%The word embedding and softmax matrices are tied \cite{Press2017UsingTO, Inan2017TyingWV}.
We partition the training data into non-overlapping blocks of 512 contiguous tokens ignoring document boundaries and train the model to predict each token from left to right \cite{Baevski2019AdaptiveIR}.
Validation and test perplexity are measured by predicting the last 256 words out of the input of 512 consecutive words to avoid evaluating tokens in the beginning with limited context (\textit{early token curse}, \citealp{shortformer}).
We generally follow the optimization method from \citet{Baevski2019AdaptiveIR}, but some hyperparameters, such as the learning rate for the \TRNN finetuning, are adjusted for better convergence than randomly initialized training.
See Appendix \ref{sec:hyper} for more details.
%and a complete list of hyperparameters.

\subsubsection{Machine Translation}
We experiment with 3 translation benchmarks: WMT14 EN-DE (4.5M train pairs, \citealp{wmt2016-findings}), WMT14 EN-FR (36M, \citealp{wmt2014-findings}), and WMT17 ZH-EN (20M, \citealp{wmt2017-findings}).
We follow the preprocessing and data splits by previous work (EN-DE: \citealp{Vaswani2017AttentionIA}; EN-FR: \citealp{Gehring2017ConvolutionalST}; EN-ZH: \citealp{Hassan2018AchievingHP}).
%, wu2018pay}).
%These datasets are all encoded into subwords by BPE \cite{sennrich-etal-2016-neural}.\footnote{We run joint BPE on all language pairs except EN-ZH.}
We use the hyperparameters of the large sized transformer \cite{Vaswani2017AttentionIA}: 6 layers, 16 attention heads, 1024 model dimensions, and 4096 hidden dimensions for both the encoder and decoder.
%Similar to the language models, all subword embedding and softmax matrices are tied.
We apply dropout with $0.3$ and label smoothing with $\varepsilon=0.1$.
Following \citet{Ott2018ScalingNM}, we use an increased batch size of approximately 460K tokens.
Each randomly initialized model is trained for 30K (60K for the large EN-FR dataset) steps using Adam with a learning rate of $5\cdot10^{-4}$ and $\beta=(0.9, 0.98)$ \cite{Kingma2014AdamAM}.
We observed that convergence of the \TRNN conversion can be achieved with 20K (40K for EN-FR) steps and a reduced learning rate of $2\cdot10^{-4}$.
We average the checkpoints from the last five epochs to obtain the final model \cite{Vaswani2017AttentionIA}.
In inference, we apply beam search with size 5 and length penalty 0.6.
Consistent with previous practice, we evaluate with tokenized BLEU \cite{Papineni2001BleuAM}.
Further details are described in Appendix \ref{sec:hyper}.
%We train with a batch size of approximately 65K tokens,
%using Adam \cite{Kingma2014AdamAM} with $\beta=(0.9, 0.98)$ and $\varepsilon=10^{-6}$.

%\smin and \smax wall-clock time speedups  (\S\ref{sec:latency}) are evaluated on the same single Nvidia V100 GPU with 16GB memory.
%We focus  on inference on a GPU machine, and leave exploration of CPU optimizations to future work.
%We note that the GPU environment works favorably for NAT models due to their parallel nature.

\begin{table}[h]
\centering
\addtolength{\tabcolsep}{-2.6pt}  
%\begin{tabular}{l@{\hspace{0.2cm}}lc?ccc|ccc?ccc|ccc}
\begin{tabular}{@{} lcccc @{}}
%& \textbf{I} & \multicolumn{2}{c}{\textbf{WMT'14}}  \\
\toprule
 & &  \multicolumn{2}{c}{ppl.} & train \\
\cmidrule(lr){3-4} %\cmidrule(lr){4-5} 
\textbf{Model}& $k$ &dev.  & test & time \\
\hline
ELU + Random Init. & 128 & 22.0 & 22.8 & 470h \\
%~/phillytools/cuda-mt-elu/experimental/orig_elu_scratch_0.0001_4321/checkpoint204.pt
%RFA,  $k$$=$$256$  & 228(rr1) &  &  \\
%RFA,  $k$$=$$128$  & 107 (1e) 108 (5e) &  &  \\
%RFA,  $k$$=$$128$ + pretrain  &  114 &    \\
%RFA + Pretrain, $k$$=$$256$  &  &  &  \\
%RFA & 64 & 20.3 &  21.1 \\
% RFA & 64 & 101
RFA + Random Init. & 32 & 20.4 & 21.3 & 512h\\
\TRNN + Random Init. & 32 & 20.1 & 20.8 & 474h\\
%Trainable MLP & 64 & 230 & \\
%RFA & 32  & 120 &  \\
\hdashline
%RFA, $k$$=$$64$  &  &  &  \\
%RFA + Pretrain, $k$$=$$64$  &  &  &  \\
%T2RNN, $k$$=$$128$  &  &  &  \\
%T2RNN, $k$$=$$128$  & blew up? &  &  \\
%T2RNN ELU & 128  & 274 &   \\
ELU + Pretrain & 128  & 21.5 & 22.2 &97h \\
%T2RNN RFA & 32 & 273 &    \\
RFA + Pretrain & 32 & 20.8 & 21.6 & 104h \\
%T2RNN Trainable MLP & 64  & 18.9 & \textbf{19.6}   \\
\TRNN + Pretrain & 32 & \textbf{19.0} & \textbf{19.6} & 98h \\
\hdashline
\TRNN 75\% + Pretrain & 32 & 17.9 &18.5  &  95h\\
\hline
%Transformer  & -- & 17.9 & 18.5 &  470h\\
Pretrained Transformer  & -- & 17.9 & 18.5 &  --\\

\citet{Baevski2019AdaptiveIR}  & -- & -- & 18.7 & --\\
\bottomrule
\end{tabular}
\caption{
%\nascomment{a fact that you seem to be downplaying is that all the pretraining models incur the cost of pretraining.  I recommend you quantify that somewhere in the text; it's not clear for example if 98h + pretraining is greater than 474h.  I suspect it is.  for clarity, you might want to consider universally labeling every system with ``pretrained'' or ``random init.'' everywhere it is mentioned; what you have now requires a little extra thought.  this issue is pervasive, e.g., it affects the legends in many of your plots}
WikiText-103 language modeling results (perplexity). Train time is measured in GPU hours. The top two rows are our reimplementations of \citet{katharopoulos-et-al-2020} and \citet{RFA}. Pretrain indicates initialization with a pretrained transformer for language modeling. \TRNN 75\% indicates a model where every fourth layer from the top is kept as the original transformer layer. Perplexity (ppl.) is measured by predicting the last 256 words out of the input of 512 consecutive words. All models use 128 head dimensions. We assume access to a pretrained transformer model and measure the finetuning time in GPU hours. }
%Conventional compound splitting is applied to WMT14 EN-DE \cite{Vaswani2017AttentionIA}.}
\label{lm_results}
\end{table}

\begin{table*}[h]
\centering
\addtolength{\tabcolsep}{-0.0pt}  
%\begin{tabular}{l@{\hspace{0.2cm}}lc?ccc|ccc?ccc|ccc}
\begin{tabular}{@{} lcccccc @{}}
%& \textbf{I} & \multicolumn{2}{c}{\textbf{WMT'14}}  \\
\toprule
%& \textbf{I} & \multicolumn{2}{c}{\textbf{WMT'14}}  \\
&  \multicolumn{2}{c}{\textbf{Feature Size $k$}} &  
\multicolumn{2}{c}{\textbf{WMT14}} &  \textbf{WMT17} & \textbf{Train Time}\\ 
\cmidrule(lr){2-3} \cmidrule(lr){4-5}  \cmidrule(lr){6-6}
% &  &   &   \multicolumn{2}{c}{{BLEU}}&   \multicolumn{2}{c?}{{Speedup}} & \multicolumn{2}{c?}{{BLEU}}& \multicolumn{2}{c}{{BLEU}} \\
\textbf{Model} & cross & causal  & EN-DE & EN-FR & ZH-EN & (GPU hours)\\
 \hline
%ELU \cite{katharopoulos-et-al-2020} & 64 & 64 & 28.4 & 391(5e)& 23.4  \\
ELU + Random Init.\ & 64 & 64 & 28.4 & * & 23.4 & 120h \\
%ELU \cite{katharopoulos-et-al-2020} & 64 & 64 & eus 226& 225 & 227 \\
RFA + Random Init.\ & 32 & 4 & 28.1 & 41.7& 23.4 &135h\\
%Trainable MLP &32  &  4 & 223 (2eeus), 401, 84 (5esea) & 218(2eeus) 399(5e)& 397 (2e) 398,400 (5e)\\
\TRNN + Random Init.\ &32 &  4 & 27.5 & 39.8 & 23.1 & 123h\\

\hdashline
ELU + Pretrain & 64 & 64 & 28.4&41.8 & \textbf{23.8} & 80h\\
%RFA \cite{RFA} & 128 & 64 & \\
%RFA + Pretraining  & 128 & 64 &  \\
RFA + Pretrain & 32 & 4 & 27.6 & 41.8 &23.2 & 90h \\
%\multirow{2}{*}{RFA + Pretraining}  & 128 & 64 &  \\
\TRNN + Pretrain & 32 & 4& \textbf{28.7} &\textbf{42.1} & \textbf{23.8} & 82h\\
\hline
%Transformer Large & -- & --& 28.9 & 42.2 & 24.2  & 120h\\
Pretrained Transformer Large & -- & --& 28.9 & 42.2 & 24.2  & --\\
\citet{Vaswani2017AttentionIA} & -- &-- &  28.4 & 41.8 & -- & -- \\
\bottomrule
\end{tabular}
\caption{Machine translation test results in BLEU scores. The top two rows are our reimplementations of \citet{katharopoulos-et-al-2020} and \citet{RFA}. Pretrain indicates initialization with a trained transformer-large model.
%similar to our \TRNN conversion.
*: diverged even when running with multiple random seeds and smaller learning rates. We assume access to a pretrained transformer model and measure the finetuning time in GPU hours.}
%ELU and RFA are our implementations of \citet{katharopoulos-et-al-2020} and \citet{RFA} with the transformer large setting.}
%Conventional compound splitting is applied to WMT14 EN-DE \cite{Vaswani2017AttentionIA}.}
\label{mt_results}
\end{table*}
\subsection{Results}
\label{sec:results}
\paragraph{Language Modeling}
Seen in Table \ref{lm_results} are language modeling results in perplexity.
We observe that \TRNN with the learnable MLP feature map outperforms the other two linear transformer models by more than 2.0 perplexity points in the pretrain setting.
Unlike the other linear transformer models, \TRNN greatly benefits from pretraining (\TRNN + Pretrain: 19.6 vs.\ \TRNN + Random Init.: 20.8 test perplexity points).
%This difference suggests that using a trainable feature map is crucial in our swap-then-finetune approach.
We attribute this advantage of \TRNN to the fact that the MLP feature map is able to learn attention patterns that are similar to those of the pretrained transformer,
as evidenced in \S\ref{sec:attn_dist}.
% in the pretraining effect 
% suggests that using a trainable feature map is crucial in our swap-then-finetune approach.
%Indeed, we will see that given the same feature size, training the parameters in the MLP feature map induces more similar attention patterns to the original transformer, compared to when the MLP parameters are frozen (\S\ref{sec:attn_dist}).
%training the MLP feature map results in more similar attention distribution (weights) to the original transformer compared to when the feature map is frozen \S\ref{sec:attn_dist}. 
%Indeed, we will see that given the same kernel size, the trainable MLP produces more similar attention distribution (weights) to the original transformer than random projections in \S\ref{sec:attn_dist}.
Notice also that the \TRNN conversion is $\sim$5x faster (measured in GPU hours) than training a model from scratch.
These results illustrate that a lightweight model can be obtained without repeating the expensive training of large-scale pretrained language models such as GPT-2 and GPT-3 \cite{gpt2, gpt3}.
\TRNN's generation speedup ($\sim$4x when producing 512 consecutive words) and memory savings are later benchmarked with varying sequence lengths.
%\nascomment{something that's not coming across in this section:  is it only because the finetuning is faster than training from scratch that we might choose T2R over the baseline?  or is there an execution time savings as well?  in MT we seem to build the argument around runtime savings.  not mentioning that here is strange!}
%
There remains a gap of 1.1 perplexity points between the \TRNN and pretrained transformer models (19.6 vs.\ 18.5).
%and further improvement of the \TRNN conversion is left for future work.
%\hao{nit: don't defer to future work too frequently until the paper gets accepted}
%\jungo{fair. deleted.}
However, the gap can be closed when every fourth layer from the top is kept as the original transformer layer and the model is finetuned in the same way (\TRNN 75\%). 
This suggests that keeping a small fraction of the quadratic attention layers can provide an effective middle ground between efficiency and accuracy.\footnote{Concurrent work \cite{attention-meets} also explores reducing the number of attention layers for efficiency.}
%Fig.\ \ref{fig:ppl-kernel} shows dev.\ ppl.\ from WikiText-103 with varying kernel sizes. 
%We see that T2RNN is particularly important to achieve low ppl.\ with small kernel sizes.
%Seen in Table \ref{lm_results} are results in perplexity. The first section shows models with linear complexity trained from random initialization.

%\nascomment{something that's not coming across in this section:  is it only because the finetuning is faster than training from scratch that we might choose T2R over the baseline?  or is there an execution time savings as well?  in MT we seem to build the argument around runtime savings.  not mentioning that here is strange!}

\paragraph{Machine Translation}
Seen in Table \ref{mt_results} are machine translation results in BLEU from various configurations.
Departing from the language modeling experiments, the \TRNN model underperforms the other two linear transformer models when initialized randomly.
However, consistent with language modeling, the \TRNN model substantially benefits from pretraining (e.g., 28.7 vs.\ 27.5 BLEU points in EN-DE).
As a result, the \TRNN model achieves similar BLEU scores to the original transformer across all language pairs.
ELU trained from the pretrained transformer yields comparable performance to \TRNN, but the feature size is much larger (64 vs.\ 32 and 64 vs.\ 4 in cross and causal attention), thus leading to increased overhead, as shown later. % in generation speed and memory overhead. 
%\yz{not supported by any results from table 2?}
Note that the \TRNN finetuning time is only moderately smaller than that of randomly initialized training here,
% in these machine translation experiments, 
but further speedup in conversion can be potentially achieved with more extensive hyperparameter tuning.\footnote{We found that the batch size could be reduced for \TRNN conversion without hurting accuracy, while randomly initialized models deteriorate with small batch sizes. This suggests that the computational cost for conversion can be much lighter than training from scratch, and \TRNN is advantageous when only a limited number of GPUs are available.}

\begin{figure}[h]
\centering
    \includegraphics[width=0.44\textwidth]{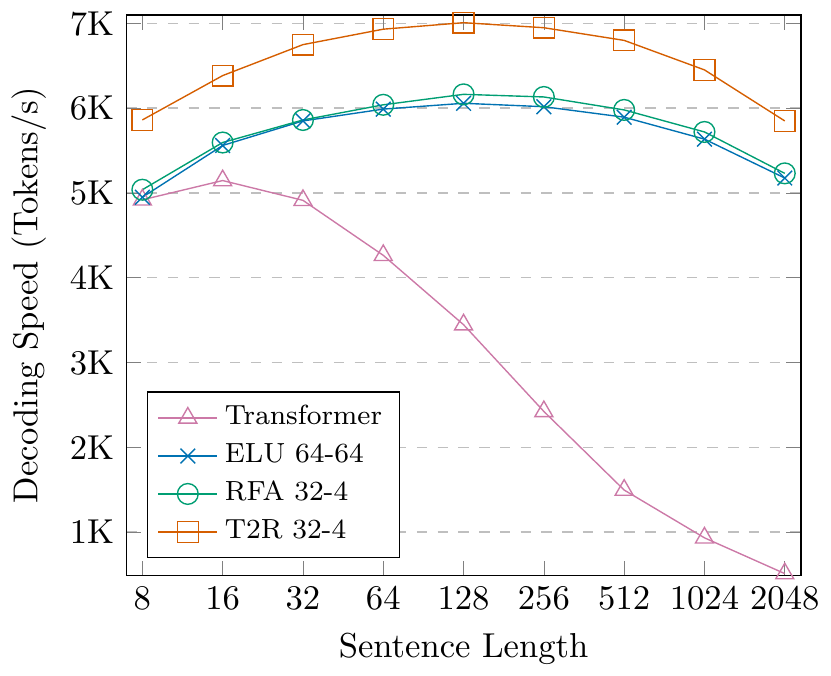}
\caption{Machine translation speed of various models. Speed is measured on a single TPU v2 accelerator with batch size 16 and beam size 1, following \citet{RFA}. 32-4 indicates the feature sizes of 32 and 4 for cross and causal attention, respectively.} 
\label{fig:mt-speed}
\end{figure}

\begin{figure}[h]
\centering
    \includegraphics[width=0.44\textwidth]{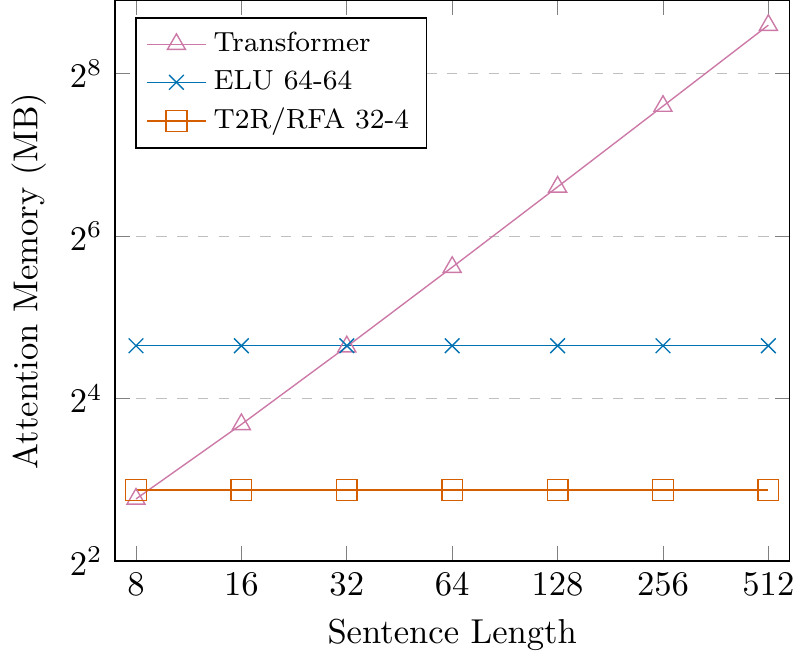}
\caption{Memory consumption from the attention computation of various machine translation models in inference with batch size 16 and beam size 1. 
% Both x and y axes are in log scale.
}
\label{fig:mt-memory}
%\vspace{-0.4cm}
\end{figure}
\paragraph{Speedup and Memory Savings in Generation}
%\hao{more details about this experiment are needed, maybe in appendix. E.g., 
%model size, the specific task (no MT has 2048 length), hardware.}
We run a conditional generation experiment to compare the decoding speed of the models in Table \ref{mt_results} (Fig.\ \ref{fig:mt-speed}).
Here we assume the input and output sequences are of the same length.
All models are tested using greedy decoding with the same batch size of 16 on a TPU v2 accelerator.\footnote{\url{https://opensource.google/projects/jax}.}
%Fig.\ \ref{fig:mt-speed} plots decoding speed with batch size 16 and beam size 1 over varying sequence lengths for machine translation.
%Here we assume that input and output sequences are of the same length.
We see that indeed the linear transformer models can generate an almost constant number of tokens per second regardless of the sequence length and outpace the transformer model dramatically as the sequence becomes longer.
The \TRNN model achieves a 15\%+ speedup over ELU and RFA due to its smaller feature sizes and faster feature mapping respectively; this confirms our analysis on \TRNN's speed advantage over them (\S\ref{sec:linear_transformers}).
Fig.\ \ref{fig:mt-memory} plots memory consumption from the attention computation during decoding for machine translation. Since the \TRNN, RFA, and ELU models compress keys and values into a $k \times d$ matrix $\mS$ and a $k$ dimensional vector $\vz$ (\S\ref{sec:convert}), the required memory at each decoding step is constant over varying sequence lengths.
It is also roughly proportional to the feature size $k$. The MLP feature map in the \TRNN model allows for small feature dimensions than the ELU feature of the head dimensions, resulting in a 70\% memory reduction.
%30\% memory overhead. 
The attention computation in the standard transformer, on the other hand, consumes memory linearly in sequence length at each decoding step because all previous key and value vectors have to be stored.
We also found a similar speedup and memory savings in unconditional generation with the \TRNN language model ($\sim$4x speedup in generating 512 consecutive words over the transformer).
%($\sim$4x speedup in generating 512 consecutive tokens over the transformer).

\section{Analysis and Ablations}
%\hao{we can highlight the conclusions upfront, especially the similarity to softmax attn}
%\jungo{Good idea. Will edit.}
We presented \TRNN, a method to convert a pretrained transformer into an efficient RNN.
In this section, we analyze our conversion approach by examining the impact of the feature size and induced attention weight distributions.
Our analysis shows that \TRNN implicitly learns attention distributions similar to the original transformer.

\begin{figure}[h]
\centering
    \includegraphics[width=0.44\textwidth]{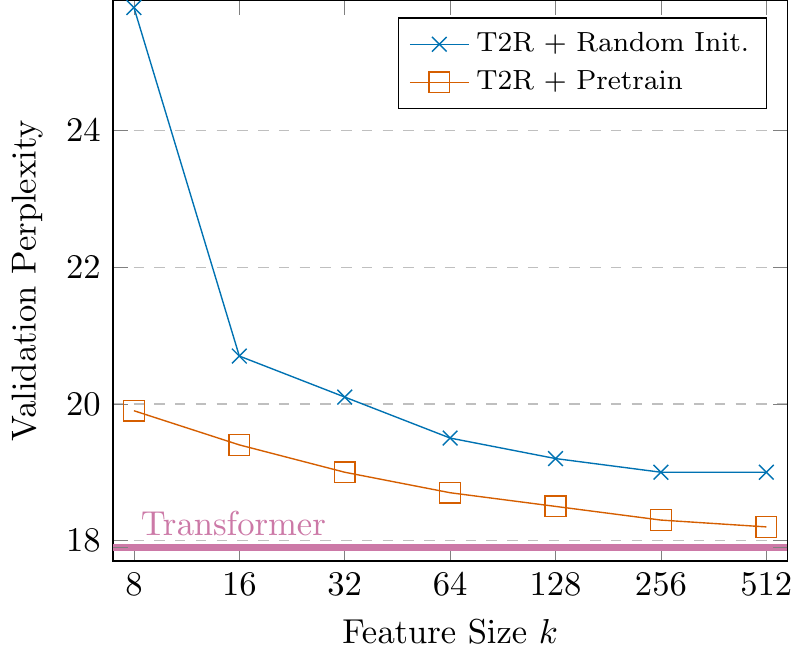}
\caption{WikiText-103 validation perplexity with varying feature sizes. %Both models use the proposed MLP feature map.
}
\label{ppl-k}
%\vspace{-0.4cm}
\end{figure}
\begin{figure}
\centering
    \includegraphics[width=0.44\textwidth]{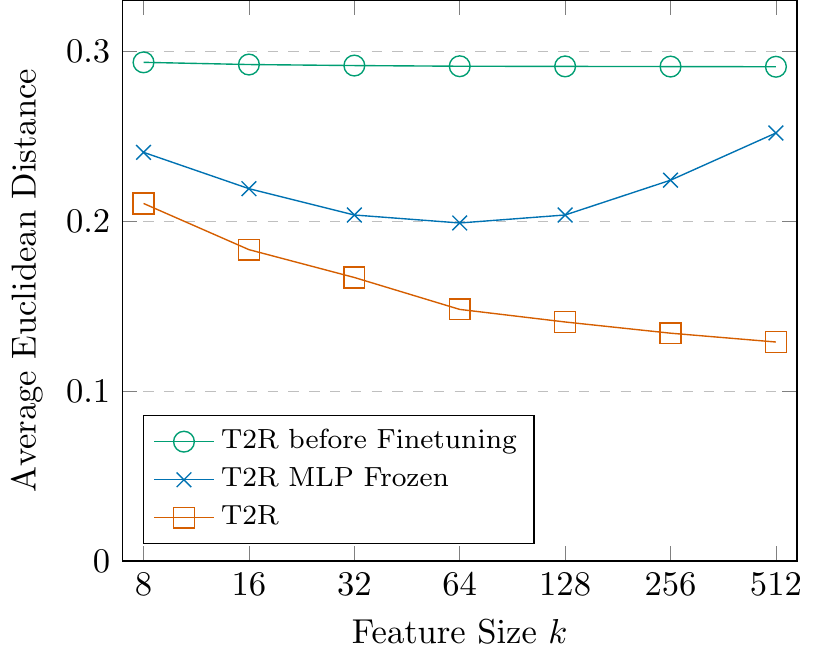}
\caption{Average Euclidean distance of \TRNN models from the transformer attention weights with varying feature sizes. The distances are computed on the Wikitext-103 validation data for predicting a word given the preceding 512 words.
All models are initialized with a pretrained transformer model.
%\nascomment{KL or JS divergence would probably be a more sensible choice for vectors that we view as distributions} 
}
\label{fig:l2-kernel}
%\vspace{-0.4cm}
\end{figure}

\paragraph{Feature Size and Pretraining}
We saw that \TRNN benefits substantially from transformer pretraining.
Fig.\ \ref{ppl-k} compares \TRNN with pretraining and random initialization in terms of the relation between the validation perplexity from WikiText-103 and the feature sizes.
We see that as the feature size (RNN state size) becomes smaller, pretraining becomes particularly important to achieve low perplexity.
Transformer pretraining achieves a Pareto improvement over random initialization in the tradeoff between efficiency (small feature size) and accuracy (low perplexity).
%\yz{if possible, I would add one dotted curves (only T2RNN) about the inference speed (tok/s) on the figure 4. People may be curious about speed vs kernel size. }

%\subsection{Sequence Length and Performance}
%In addition to standard transformer models, we compare performance with previous autoregressive transformer models with linear time and space complexity in input sequence length \cite{katharopoulos-et-al-2020, RFA}.
%As discussed in \S \ref{T2RNN}, those two methods correspond to two different untrainable feature maps $\vphi$.
%We experiment with these models in two settings: \textit{train from scratch} and \textit{T2RNN}.

\paragraph{Attention Distribution}
\label{sec:attn_dist}
%As discussed earlier, RNN models change the similarity function in the standard attention mechanism in such a way that renders a finite dimensional recurrent state. 
%Our \TRNN conversion runs a swap-then-finetune procedure; we modify the similarity function and finetune the model and the MLP parameters with the task objective.
%This means that
\TRNN is not explicitly trained to mimic the original attention distributions, and there is no guarantee that the MLP feature map approximates the exponential similarity function, unlike previous approximation approaches \cite{RFA, performer}.
Here, we analyze the properties of the attention weight distributions that are induced by finetuning.
We use the validation data from WikiText-103 and run language models to predict the next word given the input of 512 contiguous words.
We compute the attention weight distribution over the 512 words for each attention head in the model layers.

%\paragraph{Distance from Softmax}
Fig.\ \ref{fig:l2-kernel} compares the attention distributions from \TRNN in various configurations.
\TRNN MLP frozen indicates a model that is finetuned with the MLP parameters frozen.
Euclidean distances in attention distributions between the original transformer and each model are averaged across validation samples, model layers, and attention heads.\footnote{We do not consider random initialization baselines here because random initialization makes it impossible to align attention heads and layers between models.}
Comparing \TRNN before finetuning and the full \TRNN model, we see that the finetuning process induces much more similar attention distributions, and the distance diminishes as the feature size increases (and the perplexity approaches the original transformer, Fig.\ \ref{ppl-k}).
We also observed that when the MLP parameters are not trained (\TRNN MLP frozen), the distance from the original attention distributions increases.
These results suggest that finetuning of the whole network in \TRNN implicitly develops similar attention distributions to the original transformer even though the training supervision comes solely from language modeling.

\section{Further Related Work}
\label{sec:related_work}
In addition to the work we already discussed, we highlight related methods from prior work that make transformer models efficient.
\subsection{Knowledge Distillation}
Knowledge distillation \cite{Hinton2015DistillingTK} is closely related to our \TRNN conversion and uses a similar pipeline: a teacher model with large capacity is first trained and is used to generate \textit{silver} training data for a new lightweight inference model. It has been successfully applied to machine translation (e.g., \citealp{Kim2016SequenceLevelKD, Gu2017NonAutoregressiveNM}) to make generation efficient.
In particular, several prior works distill a transformer translation model to an RNN \cite{senellart-etal-2018-opennmt, kim-etal-2019-research}. 
We share the same motivation toward fast generation with light memory, but our approach differs in two ways: the original training data are used for finetuning an RNN model, and its model parameters are initialized with the ``teacher'' transformer.
%These differences have several crucial implications in practice.
Our method does not use the computationally expensive teacher model to generate new training data.
While data generation is a one-time computational cost, it becomes expensive as the teacher model size and training data increase.
Moreover, since the pretrained parameters can be directly used, conversion requires fewer GPU hours than training a brand new lightweight model from scratch (\S\ref{sec:results}).

\subsection{Efficient Transformers}
Prior work suggested many other strategies to improve efficiency in transformers, such as weight sharing and factorization \cite{universal, albert}, weight and layer pruning \cite{weightpruning,layerdrop}, quantization \cite{q8bert,quantshen}, and modifying the combination of sublayers \cite{press-etal-2020-sand,modifysublayers}.
Some of these methods present orthogonal design choices and can be integrated into our \TRNN model to gain further efficiency. 
For a more comprehensive survey, see \citet{Tay2020EfficientTA}.
Below we describe several prior works along two major strategies: compressing the attention context and sparsifying the attention patterns.
%Here we describe methods to the attention mechanism in particular.
%This work focused on improving the efficiency of the attention mechanism. 
%For a detailed overview we refer the reader to Tay et al. (2020d).
\paragraph{Attention Context Compression}
%The standard transformer architecture attends to all input tokens as context.
This strand of methods compresses the context that is attended to, thereby reducing the time and memory overhead in the attention.
RNN models that we converted pretrained transformers into compress the context into a recurrent state.
Other approaches include low rank approximation of the attention computation \cite{Wang2020LinformerSW, synthesizer} and adding a memory module that can access multiple tokens at once \cite{liuetal2018,dai-etal-2019-transformerxl,pmlr-v97-lee19d,ainslie-etal-2020-etc,raeetal2020,longformer,bigbird}.

\paragraph{Sparse Attention Patterns}
Another approach to reducing the time and memory overhead from the attention computation is to limit the tokens that are attended to by sparsifying the attention patterns.
These patterns can be set in advance or learned during training \cite{Tay2020EfficientTA}.
For example, prior works introduced fixed patterns of blockwise attention \cite{qiu-etal-2020-blockwise} and strided attention \cite{Child2019,longformer,bigbird}.
Other previous works presented methods to learn attention patterns from data \cite{sukhbaatar-etal-2019-adaptive,context-based-sparse,sinkhorn}.

It should be noted that significant modifications are necessary to apply many of these methods to autoregressive generation tasks such as language modeling and machine translation, and their empirical evaluation in these generation settings has yet to be conducted \cite{RFA}.
This work presents extensive empirical evaluation in autoregressive generation settings.
%when parallel computation is limited.

\section{Conclusion and Future Work}
We present \TRNN, a method that converts a pretrained transformer to a recurrent neural network that reduces the time and memory cost of autoregressive generation.
Our experiments in language modeling and machine translation demonstrated that our model produces an improved tradeoff between efficiency and accuracy over randomly initialized training and previous models with lightweight attention.
Our work provides further support for the claim that large-scale pretrained models can be compressed into efficient inference models that facilitate downstream applications.

\section*{Acknowledgments}
We thank Ofir Press, Bill Dolan, Lei Li, and the anonymous reviewers for their valuable feedback and discussion on this work. Nikolaos Pappas was supported by the Swiss National Science Foundation grant P400P2\_183911.

%\section*{Acknowledgments}
%We thank Ofir Press, Bill Dolan and Lei Li for their valuable feedback and discussion on this work.
\bibliography{custom}
\bibliographystyle{acl_natbib}

\appendix
\newpage
\newpage

\section{Appendix}
\subsection{Hyperparameters and Setting}
\label{sec:hyper}
All training is implemented in \texttt{fairseq} \cite{ott-etal-2019-fairseq} and run with PyTorch 1.7.1 \cite{pytorch}, 8 Telsa V100 GPUs, and CUDA 11.0.
We used mixed precision and distributed training over 8 GPUs \cite{micikevicius2018mixed, Ott2018ScalingNM}.
Apart from EN$\rightarrow$ZH where we used separate BPE operations and only tied the decoder input and output embeddings, we tie all embeddings \cite{Press2017UsingTO, Inan2017TyingWV}.
We experimented with feature sizes of [16, 32, 64] and [4, 8, 16, 32] for language modeling and machine translation respectively, and chose the smallest feature sizes that retained the development performance compared to the standard transformer. 

\subsubsection{Language Modeling}
We generally follow the optimization method from \citet{Baevski2019AdaptiveIR}.
For optimizing a model from random initialization, the learning rate is linearly warmed up from $10^{-7}$ to $1$ for the initial 16K steps and then annealed using a cosine learning rate schedule with cycles \cite{cosine}.
Each period lasts for twice the number of updates than the previous cycle, and we lower the maximum and minimum learning rates by 25\% compared to the previous cycle.
The initial minimum and maximum learning rates are $10^{-5}$ and $1$ respectively \cite{Baevski2019AdaptiveIR}.
We train the model with a batch size of about 74K tokens with a total of 286K steps \cite{Baevski2019AdaptiveIR}.
When we convert a pretrained transformer to an RNN model by finetuning, we found that we could speed up training by reducing the warm-up steps, total update steps, maximum and minimum rates, and batch size to 8K steps, 142K steps, $5\cdot10^{-6}$, $0.5$, and 25K tokens without loss in validation perplexity.
\paragraph{Randomly Initialized Training}
We generally follow the hyperparameters chosen in \citet{Baevski2019AdaptiveIR,layerdrop}.
Specifically, we list the hyperparameters in Table \ref{tab:lm-hyp} for easy replication. All other hyperparameter options are left as default values in \texttt{fairseq}.
\paragraph{Finetuning Pretrained Transformer}
Seen in Table \ref{tab:lm-ft-hyp} are the hyperparameters for finetuning a pretrained transformer to RNN models. The learning rates, the max number of updates, and the learning period length are all reduced.
\begin{table}[h]
\small
\centering
\begin{tabular}{ |l r|}
\hline
architecture & transformer\_lm\_wiki103\\
criterion & adaptive\_loss \\
tokens-per-sample & 512 \\
sample-break-mode & none \\
\# max tokens & 3072\\
dropout rate & 0.2\\
layer dropout rate & 0.2\\
decoder embed dim  & 1024\\
decoder ffn dim  & 4096\\
\# decoder attn heads & 8\\
optimizer &  nag \\
lr-scheduler &  cosine \\
lr-period-updates &  270K \\
lr-shrink & 0.75 \\
t-mult & 2 \\
max-lr & 1 \\
min-lr & 1e-9 \\
lr &  1e-4 \\
clip-norm & 0.1\\
warm-up lr & 1e-7 \\
\# warmup updates & 16K \\
\# max updates &  286K \\
\# GPUs & 8 \\
update-freq & 3\\
\hline
\end{tabular}
\caption{Language modeling hyperparameters when randomly initialized in the \texttt{fairseq} library.}
\label{tab:lm-hyp}
\end{table}
\begin{table}[h]
\small
\centering
\begin{tabular}{ |l r|}
\hline
architecture & transformer\_lm\_wiki103\\
criterion & adaptive\_loss \\
tokens-per-sample & 512 \\
sample-break-mode & none \\
\# max tokens & 3072\\
dropout rate & 0.2\\
layer dropout rate & 0.2\\
decoder embed dim  & 1024\\
decoder ffn dim  & 4096\\
\# decoder attn heads & 8\\
optimizer &  nag \\
lr-scheduler &  cosine \\
lr-period-updates &  135K \\
lr-shrink & 0.75 \\
t-mult & 2 \\
max-lr & 0.5 \\
min-lr & 1e-9 \\
lr &  5e-5 \\
clip-norm & 0.1\\
warm-up lr & 1e-7 \\
\# warmup updates & 8K \\
\# max updates &  142K \\
\# GPUs & 8 \\
update-freq & 1\\
\hline
\end{tabular}
\caption{Finetuning language modeling hyperparameters in the \texttt{fairseq} library. The learning rates are smaller than randomly initialized training.}
\label{tab:lm-ft-hyp}
\end{table}

\subsubsection{Machine Translation}
We experiment with 3 translation benchmarks: WMT14 EN-DE (4.5M train pairs, \citealp{wmt2016-findings}), WMT14 EN-FR (36M, \citealp{wmt2014-findings}), and WMT17 ZH-EN (20M, \citealp{wmt2017-findings}).
We follow the preprocessing and data splits by previous work (EN-DE: \citealp{Vaswani2017AttentionIA}; EN-FR: \citealp{Gehring2017ConvolutionalST}; EN-ZH: \citealp{Hassan2018AchievingHP, wu2018pay}).
These datasets are all encoded into subwords by BPE \cite{sennrich-etal-2016-neural}.
We run joint BPE on all language pairs except EN-ZH.
We use the hyperparameters of the large sized transformer \cite{Vaswani2017AttentionIA}: 6 layers, 16 attention heads, 1024 model dimensions, and 4096 hidden dimensions for both the encoder and decoder.
%Similar to the language models, all subword embedding and softmax matrices are tied.
We apply dropout with $0.3$, weight decay with $0.01$ and label smoothing with $\varepsilon=0.1$.
Following \citet{Ott2018ScalingNM}, we use an increased batch size of approximately 460K tokens by accumulating gradients without updating parameters.
\paragraph{Randomly Initialized Training}
We generally follow the hyperparameters chosen in \citet{Vaswani2017AttentionIA, Ott2018ScalingNM}.
Specifically, we list the hyperparameters in Table \ref{tab:mt-hyp} for easy replication. All other hyperparamter options are left as default values in \texttt{fairseq}.
The parameters from the last five epochs were averaged to obtain the final model.

\begin{table}[h]
\small
\centering
\begin{tabular}{ |l r|}
\hline
architecture & transformer\_vaswani\_en\_de\_big\\
criterion & label\_smoothed\_cross\_entropy\\
label smoothing & 0.1\\
\# max tokens & 3584 \\
dropout rate & 0.3\\
weight decay & 0.0 \\
encoder embed dim  & 1024\\
encoder ffn dim  & 4096\\
\# encoder attn heads & 16\\
\# encoder layers & 6 \\
decoder embed dim  & 1024\\
decoder ffn dim  & 4096\\
\# decoder attn heads & 16\\
\# decoder layers & 6 \\
max source positions & 1024 \\
max target positions & 1024 \\
Adam lrate& 5e-4, 3e-4 (\TRNN)*\\
Adam $\beta_1$& 0.9\\
Adam $\beta_2$& 0.98\\
lr-scheduler &  inverse square \\
warm-up lr & 1e-7 \\
\# warmup updates & 4000 \\
\# max updates &  30K, 60K (EN-FR) \\
length penalty & 0.6\\
beam size & 5\\
\# GPUs & 8 \\
update-freq & 16\\
\hline
\end{tabular}
\caption{Machine translation hyperparameters when randomly initialized in the \texttt{fairseq} library. *: we reduced the learning rate for \TRNN to avoid training divergence.}
\label{tab:mt-hyp}
\end{table}

\paragraph{Finetuning Pretrained Transformer}
Seen in Table \ref{tab:mt-ft-hyp} are the hyperparameters for finetuning a pretrained transformer to RNN models. The learning rate and the max number of updates are reduced.
The parameters from the last five epochs were again averaged to obtain the final model.
\begin{table}[h]
\small
\centering
\begin{tabular}{ |l r|}
\hline
architecture & transformer\_vaswani\_en\_de\_big\\
criterion & label\_smoothed\_cross\_entropy\\
label smoothing & 0.1\\
\# max tokens & 3584 \\
dropout rate & 0.3\\
weight decay & 0.0 \\
encoder embed dim  & 1024\\
encoder ffn dim  & 4096\\
\# encoder attn heads & 16\\
\# encoder layers & 6 \\
decoder embed dim  & 1024\\
decoder ffn dim  & 4096\\
\# decoder attn heads & 16\\
\# decoder layers & 6 \\
max source positions & 1024 \\
max target positions & 1024 \\
Adam lrate& 2e-4 \\
Adam $\beta_1$& 0.9\\
Adam $\beta_2$& 0.98\\
lr-scheduler &  inverse square \\
warm-up lr & 1e-7 \\
\# warmup updates & 4000 \\
\# max updates &  20K, 40K (EN-FR) \\
length penalty & 0.6\\
beam size & 5\\
\# GPUs & 8 \\
update-freq & 16\\
\hline
\end{tabular}
\caption{Finetuning machine translation hyperparameters. The learning rate is smaller than randomly initialized training.}
\label{tab:mt-ft-hyp}
\end{table}

\subsection{Attention Distribution}
\begin{figure}[h]
\centering
    \includegraphics[width=0.49\textwidth]{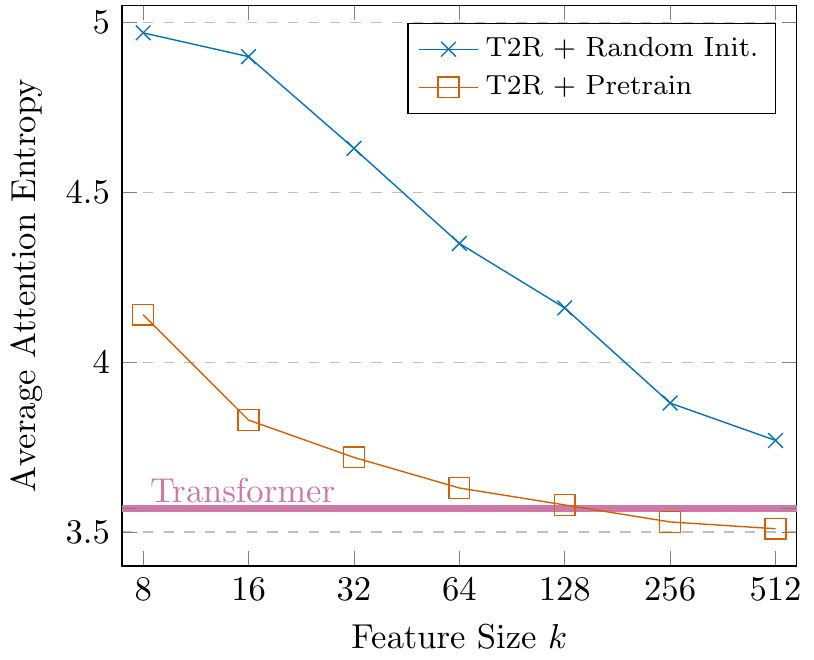}
\caption{Average entropy of the attention weights. They are computed on the Wikitext-103 validation data for predicting a word given the preceding 512 words.}
\label{fig:entropy-kernel}
%\vspace{-0.4cm}
\end{figure}

\paragraph{Peakiness of Attention}
Fig.\ \ref{fig:entropy-kernel} plots the average entropy of the \TRNN models with and without pretraining.
%\footnote{Attention distribution entropy for RFA is not computed here since the Monte Carlos approximation of the Gaussian kernel \cite{rahimi2007} in RFA can result in negative weights.}
Entropy is averaged across validation samples, layers, and attention heads.
Comparing Figs.\ \ref{ppl-k} and \ref{fig:entropy-kernel}, we see that there is strong correlation between validation perplexity and entropy.
The entropy decreases (and thus the attention distribution gets peakier) when a large feature size is used or the transformer pretraining is applied.
This observation hints at potential future improvement of linear transformer models by introducing an inductive bias towards peaky attention distributions.

\end{document}